\def\year{2020}\relax
\documentclass[letterpaper]{article} % DO NOT CHANGE THIS
\usepackage{aaai20}  % DO NOT CHANGE THIS
\usepackage{times}  % DO NOT CHANGE THIS
\usepackage{helvet} % DO NOT CHANGE THIS
\usepackage{courier}  % DO NOT CHANGE THIS
\usepackage[hyphens]{url}  % DO NOT CHANGE THIS
\usepackage{graphicx} % DO NOT CHANGE THIS
\usepackage{graphicx}  % DO NOT CHANGE THIS
\usepackage{epsfig}
\usepackage{amsmath}
\usepackage{amssymb}

\usepackage{comment}
\usepackage{multirow}
\usepackage{amsthm}
\usepackage{color}

\usepackage{enumitem}
\usepackage{subfigure}
\usepackage{graphicx}
\usepackage{xcolor}
\usepackage{url}
\usepackage{booktabs}       % professional-quality tables
\usepackage{threeparttable}
\usepackage{multirow}

\usepackage{algorithmic}
\usepackage[ruled,linesnumbered]{algorithm2e}
\usepackage{bm}
\usepackage{tikz}
\usepackage{adjustbox}      
\usepackage{filecontents}
\usepackage{lipsum,booktabs}
\usepackage{tablefootnote}

\usepackage{float}
\restylefloat{table}

\usepackage{latexsym}
\usepackage{amsmath}
\usepackage{amsfonts}
\usepackage{amssymb}

\renewcommand{\|}{\textrm{~}\arrowvert\textrm{~}}

\newcommand{\by}{\sim}

\allowdisplaybreaks[3]

\newcommand{\beq}{\begin{equation}}
\newcommand{\eeq}{\end{equation}}
\newcommand{\beqs}{\begin{eqnarray}}
\newcommand{\eeqs}{\end{eqnarray}}
\newcommand{\barr}{\begin{array}}
      \newcommand{\earr}{\end{array}}

\newcommand{\Av}{{\boldsymbol A}}

\newcommand{\Dv}{{\boldsymbol D}}

\newcommand{\Hv}{{\boldsymbol H}}
\newcommand{\Iv}{{\boldsymbol I}}

\newcommand{\Wv}{{\boldsymbol W}}
\newcommand{\xv}{{\boldsymbol x}}
\newcommand{\yv}{{\boldsymbol y}}
\newcommand{\Xv}{{\boldsymbol X}}

\newcommand{\zv}{{\boldsymbol z}}

\newcommand{\alphav}{{\boldsymbol \alpha}}
\newcommand{\betav}{{\boldsymbol \beta}}

\newcommand{\epsilonv}{{\boldsymbol \epsilon}}

\newcommand{\muv}{{\boldsymbol \mu}}

\newcommand{\sigmav}{{\boldsymbol \sigma}}
\newcommand{\Sigmav}{{\boldsymbol \Sigma}}

\newcommand{\Ncal}{\mathcal{N}}

\title{Graph-Driven Generative Models for Heterogeneous Multi-Task Learning}
\author{
Wenlin Wang\textsuperscript{\rm 1},
Hongteng Xu\textsuperscript{\rm 2},
Zhe Gan\textsuperscript{\rm 3},
Bai Li\textsuperscript{\rm 1}, 
Guoyin Wang\textsuperscript{\rm 1}, \\
\large{\textbf{Liqun Chen}\textsuperscript{\rm 1}, 
\textbf{Qian Yang}\textsuperscript{\rm 1}, 
\textbf{Wenqi Wang}\textsuperscript{\rm 4}, 
\textbf{Lawrence Carin}\textsuperscript{\rm 1}}\\
\textsuperscript{\rm 1}Duke University,
\textsuperscript{\rm 2}Infinia ML, Inc, 
\textsuperscript{\rm 3}Microsoft Dynamics 365 AI Research,
\textsuperscript{\rm 4}Facebook, Inc\\
wenlin.wang@duke.edu
}
 
\begin{document}

\maketitle

%-------------------------------------------------
\begin{abstract}
We propose a novel graph-driven generative model, that unifies multiple heterogeneous learning tasks into the same framework. 
The proposed model is based on the fact that heterogeneous learning tasks, which correspond to different generative processes, often rely on data with a shared graph structure. 
Accordingly, our model combines a graph convolutional network (GCN) with multiple variational autoencoders, thus embedding the nodes of the graph ({\em i.e.}, samples for the tasks) in a uniform manner while specializing their organization and usage to different tasks.
With a focus on healthcare applications (tasks), including clinical topic modeling, procedure recommendation and admission-type prediction, we demonstrate that our method successfully leverages information across different tasks, boosting performance in all tasks and outperforming existing state-of-the-art approaches. 
\end{abstract}

%-------------------------------------------------
\section{Introduction}
Multi-task learning aims to {\em jointly} solve different learning tasks, while leveraging appropriate information sharing across all tasks \cite{thrun1996learning,caruana1997multitask}.
It has been shown that learning under a multi-task setting usually yields enhanced performance relative to separately building single-task models~\cite{sermanet2013overfeat,hashimoto2016joint,ruder2017overview}.
However, multi-task learning has primarily been considered for \emph{homogeneous} tasks that share the same objective ({\em e.g.}, the same set of labels)~\cite{baxter1997bayesian,bakker2003task,yu2005learning,luo2015multi}. 
Real-world tasks are often \emph{heterogeneous}~\cite{jin2014multi}, meaning that each task potentially has a different objective and relies on complicated, often unobserved interactions.
Examples of tasks having different objectives include classification, regression, recommendation {\em etc}. 

From the perspective of generative models, heterogeneous tasks often correspond to distinct generative processes.
This implies that traditional generative multi-task learning methods~\cite{baxter1997bayesian,bakker2003task,yu2005learning,zhang2008flexible}, which often generalize a single class of generative model to multiple tasks, are not appropriate. 
Under these circumstances, a new mechanism is required to leverage relationships across the entities from different tasks.

To overcome the aforementioned challenges, we propose a graph-driven generative model to learn heterogeneous tasks in a unified framework. 
Taking advantage of the graph structure that commonly appears in many real-world data, the proposed model treats feature views, entities and their relationships as nodes and edges in a graph, and formulates learning heterogeneous tasks as instantiating different sub-graphs from the global data graph.
Specifically, a sub-graph contains the feature views and the entities related to a task and their interactions. 
Both the feature views and the interactions can be reused across all tasks while the representation of the entities are specialized for the task. 
We combine a shared graph convolutional network (GCN)~\cite{kipf2016semi} with multiple variational autoencoders (VAEs)~\cite{kingma2013auto}.
The GCN serves as a generator of latent representations for the sub-graphs, while the VAEs are specified to address the different tasks.
The model is then optimized jointly over the objectives for all tasks to encourage the GCN to produce representations that can be used simultaneously by all of them.

In health care, our motivating example, ICD (International Statistical Classification of Diseases) codes for diseases and procedures can be used as a source of information for multiple tasks, {\em e.g.}, modeling clinical topics of admissions, recommending procedures according to diseases and predicting admission types.
These three tasks require the capture of clinical relationships among ICD codes and admissions. 
For a given admission, it is associated with a set of disease and procedure codes ({\em i.e.}, feature views). 
\begin{figure*}[t!]
    \centering
    \includegraphics[width=0.85\textwidth]{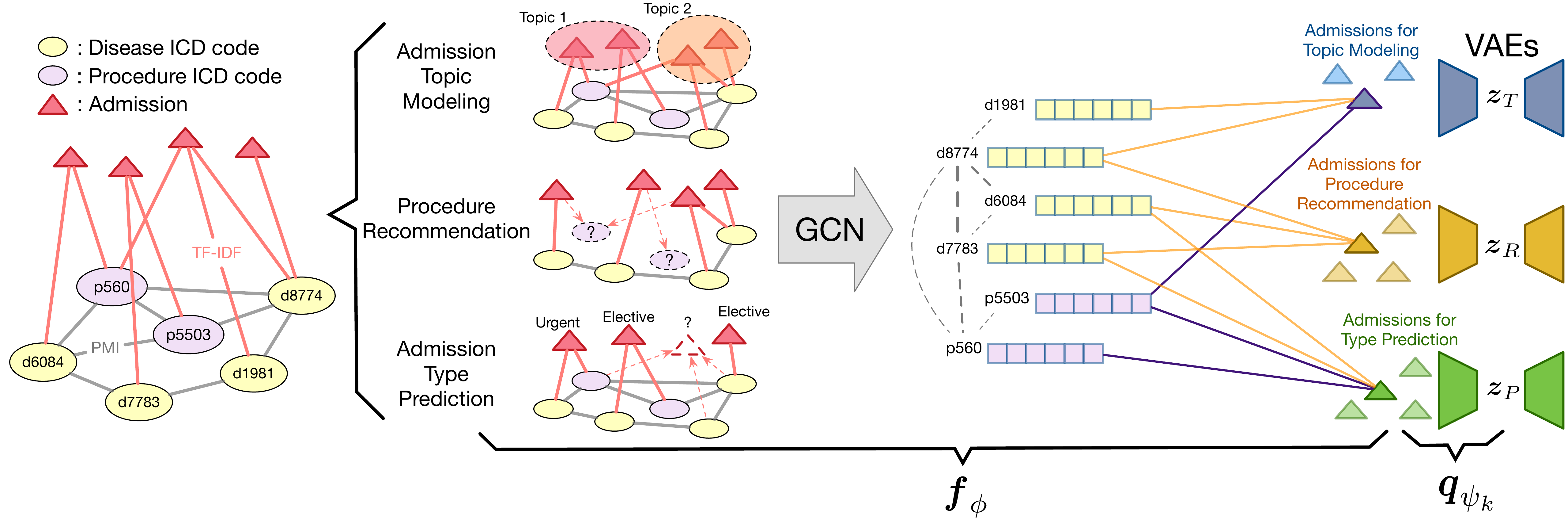}
    \caption{Illustration of the proposed model for healthcare tasks. 
    Each task operates on a different sub-graph from the admission graph. 
    The shared GCN ($f_{\phi}$) learns embeddings for ICD codes and admissions, and the embeddings pass through task-specific VAEs accordingly.
    }
    \label{fig:scheme}
\end{figure*}
However, the admission has to be organized with different views ({\em i.e.}, specialized entities) for tasks with different objectives. 
For instance, topic modeling is an unsupervised task needing procedures and diseases, admission-type prediction is a supervised task also using procedures and diseases, and procedure recommendation is a supervised task that only uses disease codes. 
In the context of our work, ICD codes and hospital admissions constitute a graph as shown in Figure~\ref{fig:scheme}. 
The edges between ICD codes and those between ICD codes and admissions are quantified according to their coherency. 
The ICD codes are embedded during training, which are used to specialize the embeddings of admissions for different tasks. 
At test time, the GCN is used to represent sub-graphs, {\em i.e.}, collections of shared ICD codes, specialized admissions and their interactions, that feed into different task-specific VAEs.
We test our model on the three tasks described above.
Experimental results show that the jointly learned representation for the admission graph indeed improves the performance of all tasks relative to the individual task model.

%------------------------------------------------------------------------
\section{Proposed Model}
To solve heterogeneous multi-task learning from a generative model perspective, a natural solution is to model multiple generative processes, one for each task.
In particular, given $K$ tasks, each task $k$ is associated with training data $(\xv_k, \yv_k)$, where $\yv_k$ represents the target variable, and $\xv_k$ represents the variable associated with $\yv_k$.
We propose using $K$ sets of VAEs~\cite{kingma2013auto} for modeling $\{\yv_k\}_{k=1}^K$ in terms of latent variables $\{\zv_k\}_{k=1}^K$, where each $\zv_k$ is inferred from $\xv_k$ using a task-specific inference network.
Note that here the term VAE is used loosely in the sense that $\yv_k$ and $\xv_k$ need not to be the same.
The generative processes are defined as
\begin{align}
    \yv_k \sim p_{\theta_k}(\yv_k|\zv_k)\,, \quad \zv_k \sim p (\zv_k)\,,\quad k = 1,\ldots,K.
\end{align}
with corresponding inference networks specified as
\begin{align}
    \zv_k \sim q_{\psi_k} (\zv_k|f_{\phi}(\xv_k))\,,\quad k = 1,\ldots,K.
\end{align}
For the $k$-th task, $p_{\theta_k}(\cdot)$ represents a generative model (\emph{i.e.}, a stochastic decoder) with parameters $\theta_k$, and $p(\zv_k)$ is the prior distribution for latent code $\zv_k$.
The corresponding inference network for $\zv_k$ consists of two parts: ($i$) a deterministic encoder $f_{\phi}(\cdot)$ shared across all tasks to encode each $\xv_k$ into $\hat{\xv}_k=f_{\phi}(\xv_k)$ independently; and ($ii$) a stochastic encoder with parameters $\psi_k$ to stochastically map $\hat{\xv}_k$ into latent code $\zv_k$.
The distribution $q_{\psi_k} (\zv_k|f_{\phi}(\xv_k))$ serves as an approximation to the unknown true posterior $p(\zv_k|\yv_k)$.
Note that since $\{\yv_k\}_{k=1}^K$ are in general associated with heterogeneous tasks, they may represent different types of information.
For example, they can be labels for classification and bag-of-words for topic modeling.
Motivated by the intuition that real-world tasks are likely to be latently related with each other, using a shared representation $f_{\phi}(\cdot)$ can be beneficial as a means to consolidate information in a way that allows tasks to  leverage information from each other.

In likelihood-based learning, the goal for heterogeneous multi-task learning is to maximize the empirical expectation of the log-likelihood $\frac{1}{K}\sum_{k=1}^K \log( p(\yv_k) )$, with respect to the data provided for each task. 
Since the marginal likelihood $p(\yv_k)$ rarely has a closed-form expression, VAE seeks to maximize the following evidence lower bound (ELBO), which bounds the marginal log-likelihood from below
\begin{align}\label{eqn:objective}\small
    \mathcal{L}(\theta_{1:K},\psi_{1:K},\phi) &= \sideset{}{_{k}}\sum \Big[ \mathbb{E}_{q_{\psi_k}(\zv_k|f_{\phi}(\xv_k))}[\log p_{\theta_k}(\yv_k|\zv_k)] \nonumber \\ 
    &- \text{KL}(q_{\psi_k}(\zv_k|f_{\phi}(\xv_k)) \parallel p(\zv_k)) \Big] \,.
\end{align}
However, for heterogeneous tasks, features are often organized in different views and the interactions between observed entities can as well be different. 
As a result, it is challenging to find a common $f_{\phi}(\cdot)$ for the $\{\xv_k\}_{k=1}^K$ with incompatible formats or even in incomparable data spaces.

Fortunately, such data can often be modeled as a data graph, whose nodes correspond to the entities appearing in different tasks and edges capturing their complex interactions. 
Accordingly, different tasks re-organize the graph and leverage its information from shared but different views. 
Specifically, we represent a data graph as $G(\mathcal{V}, \mathcal{X}, \Av)$, where   
$\mathcal{V}=\{v_1,v_2,...\}$ is the set of nodes corresponding to the observed entities,  
$\Av\in\mathbb{R}^{|\mathcal{V}|\times |\mathcal{V}|}$ is the adjacency matrix of the graph,  
and $\mathcal{X}=\cup_{k=1}^{K}\mathcal{X}_k$ is a union of (trainable) feature sets. 
For the $k$-th task, $\mathcal{X}_k=\{\xv_{v,k}\}_{v\in\mathcal{V}_k}$ is its feature set, where $\mathcal{V}_k\subset \mathcal{V}$ contains the nodes related to the task and  $\xv_{v,k}$ is the feature of the node $v$ in task $k$. 
Based on $\mathcal{X}_k$, the observations of the $k$-th task correspond to a sub-graph from $G$, {\em i.e.}, $G_k=G(\mathcal{V}_k, \mathcal{X}_k, \Av_k)$, where $\Av_k = \Av(\mathcal{V}_k, \mathcal{V}_k)$ selects rows and columns from $\Av$. 
In such a situation, instead of finding a unified inference network for each individual observation in different tasks, for the sub-graphs we define an inference network based on a graph convolutional network (GCN)~\cite{kipf2016semi}, {\em i.e.}, implementing $f_{\phi}(\cdot)$ in (\ref{eqn:objective}) as a GCN with parameter $\phi$ and thus $\zv_k\sim q_{\psi_k}(\zv_k|\text{GCN}_{\phi}(G_k))$, hence a large portion of the parameters of the inference network are shared among tasks.

The independent generative processes with a shared GCN-based inference network match with the nature of heterogeneous tasks. 
In particular, the sub-graphs in different tasks are derived from the same data graph with partially shared nodes and edges, enabling joint learning of latent variables through the shared inference network. 
Then, the inferred latent variables pass through different generative models under the guidance of different tasks.
In the next section, we will show that this model is suitable for challenging healthcare tasks.

%------------------------------------------------------------------------
\section{Typical Specification for Healthcare Tasks}
\subsection{Observations, tasks, and proposed data graph}
To demonstrate the feasibility of our model, we describe a specification to solve tasks associated with hospital admissions. 
Let $\mathcal{V}^d = \{v_1^d, v_2^d,\ldots\}$ and $\mathcal{V}^p = \{v_1^p, v_2^p,\ldots\}$ denote the set of disease and procedure ICD codes, respectively, {\em i.e.},  each component $v_i^d\in \mathcal{V}^d$ represents a specific disease and each $v_i^p\in \mathcal{V}^p$ represents a specific procedure. 
Suppose we observe $N$ hospital admissions, denoted as $\mathcal{V}^a=\{v_1^a, v_2^a,\ldots, v_N^a\}$. 
Each $v_n^a\in\mathcal{V}^a$ is associated with some ICD codes and a label representing its type, {\em i.e.}, $\{\mathcal{V}_n^d, \mathcal{V}_n^p, c_n\}$ for $n=1,\ldots,N$, where $\mathcal{V}_n^d\subseteq \mathcal{V}^d$, $\mathcal{V}_n^p\subseteq \mathcal{V}^p$ and $c_n\in\mathcal{C}$ is an element in the set of admission types $\mathcal{C}$. 
Based on these observations, we consider three healthcare tasks: $i$) clinically-interpretable topic modeling of admissions; $ii$) procedure recommendation; and $iii$) admission-type prediction.

As illustrated in Figure~\ref{fig:scheme}, the observations above can be represented as an admission graph $G(\mathcal{V},\mathcal{X},\Av)$, where the node set $\mathcal{V}=\mathcal{V}^d\cup\mathcal{V}^p\cup\mathcal{V}^a$ and $\Av\in\mathbb{R}^{|\mathcal{V}|\times |\mathcal{V}|}$ is the adjacency matrix. 
The union of feature sets $\mathcal{X}=\mathcal{X}^d\cup\mathcal{X}^p\cup\mathcal{X}^a$, where $\mathcal{X}^d=\{\xv_{v}\}_{v\in\mathcal{V}^d}$ and $\mathcal{X}^p=\{\xv_v\}_{v\in\mathcal{V}^p}$ contain trainable vector embeddings of ICD codes for diseases and procedures, respectively. 
These embeddings are reused for different tasks.
$\mathcal{X}^a=\cup_{k=1}^{K}\mathcal{X}^a_k$, where $\mathcal{X}^a_k=\{\xv_v^k\}_{v\in\mathcal{V}^a}$ contains the embeddings of admissions for different tasks. 
Specifically, for each admission $v_n^a$, its embedding in the $k$-th task is derived from the aggregation of the embeddings of ICD codes, {\em i.e.},  $\xv_{v_n^a}^k=\text{MaxPooling}(\{\xv_v\}_{v\in \mathcal{V}_n^k})$, where $\mathcal{V}_n^k$ is the set of the ICD codes associated with task $k$. 
For topic modeling and admission-type prediction, $\mathcal{V}_n^k=\mathcal{V}_n^d\cup \mathcal{V}_n^p$, while for procedure recommendation, in which the procedure codes are unavailable, $\mathcal{V}_n^k=\mathcal{V}_n^d$. 
Given this admission graph $G$, the three healthcare tasks correspond to different sub-graphs $G(\mathcal{V}_k,\mathcal{X}_k,\Av_k)$, which yield a typical heterogeneous scenario.
Table~\ref{tab:Variables} highlights their differences on target variables and sub-graphs. Although the sub-graphs specialize the information of admission nodes, they reuse the representations of ICD code nodes and the edges in $G$. 

\subsection{Construction of edges}
Inspired by existing research~\cite{matveeva2006document,chen2013alternative,rekabsaz2017toward,yao2018graph}, we enrich the representation power of our model with the meaningful population statistics, considering two types of edges in our adjacency matrix. 

\textbf{Edges between ICD codes.} ICD codes appear coherently in many admissions, {\em e.g.}, diabetes and its comorbidities like cardiovascular disease. 
Accordingly, edges between ICD codes with high coherency should be weighted heavily. 
Based on this principle, we apply point-wise mutual information (PMI), which is a commonly-used similarity measurement in various NLP tasks~\cite{levy2014neural,arora2016latent,newman2010automatic,mimno2011optimizing,ogura2013text}, as the weight between each pair of ICD codes. 
Formally, for each pair of ICD codes, we evaluate their PMI as
\begin{align}
\text{PMI}(i,j) = \log p_{ij} -\log(p_i p_j), \ \text{for} \ i,j\in\mathcal{V}^d\cup\mathcal{V}^p, 
\end{align}
where 
$p_{ij} = \frac{|\{v_n^a | i,j\in \mathcal{V}_n^d\cup \mathcal{V}_n^p\}|}{N}$ and $p_i = \frac{|\{v_n^a | i\in \mathcal{V}_n^d\cup \mathcal{V}_n^p\}|}{N}$. 
Positive PMI values indicate that the ICD codes in the pair are highly-correlated with each other. Conversely, negative PMI values imply weak correlation. 
Therefore, we only consider positive PMI values as the weights of edges. 

\textbf{Edges between ICD codes and admissions.} Analogous with the relationship between words and documents, we weight the edge between ICD codes and admissions with the help of the term frequency-inverse document frequency (TF-IDF)\footnote{\url{https://en.wikipedia.org/wiki/Tf–idf}}. 
The term frequency (TF) is the normalized version of the number of times an ICD code appears in an admission, and the inverse document frequency (IDF) is the log-scaled inverse fraction of the number of admissions that contain the ICD code. 
The TF-IDF is the element-wise multiplication of TF and IDF, which defines how important an ICD code in an admission~\cite{onan2016ensemble,shen2018baseline}. 

Summarizing the above, elements $a_{ij}$ in the adjacency matrix $\Av$ are represented as
\begin{eqnarray}
{\small
\begin{aligned}
a_{ij}=
\begin{cases}
1,                   & i=j\\
\text{PMI}(i, j),    & i, j\in\mathcal{V}^d\cup\mathcal{V}^p~\text{and}~\text{PMI}(i, j) > 0\\
\text{TF-IDF}(i, j), & i\in\mathcal{V}^a,~j\in\mathcal{V}^d\cup\mathcal{V}^p\\
0,                   & \text{otherwise}
\end{cases} .
\end{aligned} }
\end{eqnarray}

\subsection{Graph-driven VAEs for different tasks}
Focusing on the three tasks mentioned above, we specify our model as graph-driven variational autoencoder (GD-VAE). 
Specifically, GD-VAE consist of: $i$) a GCN-based inference network that is shared by all the three tasks, and $ii$) three specialized generative networks that account for different sets of observations corresponding to the three tasks.

\begin{table*}[t]
	\centering
	\caption{\small{Illustration of the differences between the three healthcare tasks.}}\label{tab:Variables}
	\begin{small}
	\begin{tabular}{c|cl|c}
	\hline\hline
	    \multirow{2}{*}{Task} &\multicolumn{2}{c|}{$G_k$} &\multirow{2}{*}{$\yv_k$}\\ \cline{2-3}
	    &$\mathcal{V}_k$ &$\qquad x_{v_n^a}^k$ in $\mathcal{X}_k$ &\\ \hline
		Topic Modeling
		&$\mathcal{V}$  
		&$\text{MaxPooling}(\{\xv_v\}_{v\in \mathcal{V}_n^d\cup \mathcal{V}_n^p})$
		&Bi-term ICD codes\\
		Procedure Recommendation
		&$\mathcal{V}^d\cup\mathcal{V}^a$  
		&$\text{MaxPooling}(\{\xv_v\}_{v\in \mathcal{V}_n^d})$
		&List of procedures\\
		Admission-type Prediction
		&$\mathcal{V}$  
		&$\text{MaxPooling}(\{\xv_v\}_{v\in \mathcal{V}_n^d\cup \mathcal{V}_n^p})$
		&Admission type, $c\in\mathcal{C}$\\
\hline\hline
	\end{tabular}
    \end{small}
\end{table*}

\textbf{Topic modeling of admissions} \label{sec:topic_model}
In the context of topic modeling, each ICD code can be considered as a \emph{word} or {\em token}, while each admission corresponds to a \emph{document}, {\em i.e.}, a collection of ICD codes. 
However, patient admissions exhibit extreme-sparsity issues in the sense that a very small set of codes are associated with each admission.
% in the scenario of short ICD codes and rare observation (maximum once in an admission). 
Classic topic models, such as LDA~\cite{blei2003latent} and Neural Topic Model~\cite{miao2017discovering}, can therefore be not appropriate in this case. 
To circumvent this problem, inspired by~\cite{yan2013biterm}, instead of modeling a bag-of-ICD-codes for a single admission, we aim to model bi-term collections, and then aggregate all the unordered ICD code pairs (bi-terms) from several admissions together as one document. 
% We assume the admission as a mixture of topics, and  Then, 
The generative process of our proposed Neural Bi-term Topic Model (NBTM) is described as follows:
{\small
\begin{align}
    \zv_T \sim \text{Dir}(\alphav), 
    \quad l\sim \text{Multi}(1, \zv_T), \quad \yv_T \sim \text{Multi}(2, \betav_l) \,,
\end{align} }
where $\yv_T$ is the bi-term variable and its instance is $\{v_i, v_j\}$, where $v_i, v_j$ are two ICD codes. $\zv_T$ is the topic distribution. $\alphav$ is the hyper-parameter of the Dirichlet prior; a vector with length $L$, where $L$ is the number of topics.
$\betav = \{\betav_l\in \mathbb{R}^{|\mathcal{V}^d| + |\mathcal{V}^p|}\}_{l=1}^L$ are trainable parameters, each representing a learned topic, \emph{i.e.}, the distribution over ICD codes.
The marginal likelihood for the entire admission corpus $\yv_T$ can be written as
\begin{align}\label{Eq:NBTM}\small
p(\yv_T) &= \int_{\zv_T} p(\zv_T) \sideset{}{_{i,j}}\prod\sideset{}{_l}\sum p(v_i, v_j | \betav_l)p(l |\zv_T) d\zv_T ~\nonumber \\
&= \int_{\zv_T} p(\zv_T) \sideset{}{_{i,j}}\prod p(v_i, v_j| \betav, \zv_T) d\zv_T\,.
\end{align}
The Dirichlet prior is known to be essential for generating interpretable topics~\cite{wallach2009rethinking}.
However, it can be rarely applied to VAE directly, since no effective re-parameterization trick that can be adopted for the Dirichlet distribution.
Fortunately, the Dirichlet distribution can be approximated with a logistic normal and a softmax formulation by Laplace approximation~\cite{hennig2012kernel}.
When the number of topics $L$ is large, the Dirichlet distribution can be approximated with a multivariate logistic normal~\cite{srivastava2017autoencoding} with the $i$-th element of its mean $\muv_T$ and diagonal covariance matrix $\Sigmav_T$ as follows:
{
\begin{align}
    \mu_{i}^T &= \log \alpha_i - \frac{1}{L} \sideset{}{_{j=1}^L}\sum\log \alpha_j , \nonumber \\
    \Sigma_{ii}^T &= \frac{1}{\alpha_i} (1 - \frac{2}{L}) + \frac{1}{L^2}\sideset{}{_{j=1}^L}\sum \frac{1}{\alpha_j}\,. \label{Eq:DirichletApprox}
\end{align} }
Under such an approximation, a topic distribution can be readily inferred by applying re-parameterization trick, sampling $\epsilonv\sim \mathcal{N}(\boldsymbol{0}, \Iv)$ and inferring $\zv_T$ via $\zv_T = \text{softmax}(\muv_T + \Sigmav_T^{1/2}\epsilonv)$. 

\noindent\textbf{Procedure recommendation} \label{sec:recommendation}
In this task, for an admission, we aim to predict the set of procedures $\yv_R$ for a set of diseases. 
Inspired by~\cite{liang2018variational}, we consider the following generative process for modeling admission procedures:
{\small
\begin{align}
    \zv_R\sim \Ncal(\boldsymbol{0}, \Iv), \quad \pi_R\propto \exp\{g (\zv_R)\}, \quad \yv_R\sim \text{Multi}(M, \pi_R)\,,
\end{align} }
where $\yv_R$ is $|\mathcal{V}^p|$-dimensional variable and its instance is a list of $M$ recommended procedures.
$g(\cdot)$ is a multi-layer perceptron (MLP). 
The output of this function is normalized to be a probability distribution over procedures, {\em i.e.}, $\pi_R\in \Delta^{|\mathcal{V}^p|}$, where $\Delta$ denotes a simplex.
Then we derive procedures for the given admission by sampling $M$ times from a multinomial distribution with parameter $\pi_R$.   

\noindent\textbf{Admission-type prediction} %\label{sec:prediction}
Given an admission, the goal is to predict the admission type given both its diseases and procedures. 
We consider the following generative process for modeling admission types:
{\small
\begin{align}
    \zv_P\sim \mathcal{N}(\boldsymbol 0, \Iv), \quad \pi_P\propto \exp\{h (\zv_P)\}, \quad 
    \yv_P \sim \text{Multi}(1, \pi_P)\,,
\end{align}}
where $\yv_P$ is a variable and its instance corresponds to an admission type in the set $\mathcal{C}$.
$h(\cdot)$ is another MLP, whose output is normalized to be a distribution over admission types, {\em i.e.}, $\pi_P\in\Delta^{|\mathcal{C}|}$.
Finally, the instance of $\yv_R$ (the type of the given admission) is sampled once from a multinomial distribution with parameter $\pi_P$. 

\noindent\textbf{Inference with a shared GCN}
The proposed model unifies three tasks via sharing a common GCN-based inference network. 
Specifically, the posteriors of the three latent variables are
{\small
\begin{align}
    \zv_T\sim \mathcal{LN}(\muv_T, \Sigmav_T), \
    \zv_R\sim \mathcal{N}(\muv_R, \Sigmav_R), \
    \zv_P\sim \mathcal{N}(\muv_P, \Sigmav_P),
\end{align} }
where $\text{diag}(\sigmav_k) = \Sigmav_k^{1/2}$, $\text{diag}(\cdot)$ represents a diagonal matrix, and $[\muv_k;\log\sigmav_k] = \text{MLP}_{\psi_{k}}(\text{GCN}_{\phi}(G_{k}))$ for $k \in \{P, R, T\}$. 

Let $\theta_T, \theta_R, \theta_P$ denote the parameters of the generative networks for topic modeling, procedure recommendation and admission-type prediction, respectively. 
In summary, all the parameters $\{\theta_T, \theta_R, \theta_P, \psi_P, \psi_R, \psi_T, \phi\}$ are optimized jointly via maximizing (\ref{eqn:objective}).

%------------------------------------------------------------------------
\section{Related Work}
\textbf{Multi-task learning}
Early multi-task learning methods learn a shared latent representation~\cite{thrun1996learning,caruana1997multitask,baxter1997bayesian}, or impose structural constraints on the shared features for different tasks~\cite{ando2005framework,chen2009convex}. 
The work in~\cite{he2011graphbased} proposed a graph-based framework leveraging information across multiple tasks and multiple feature views.
Following this work, the methods in~\cite{zhang2012inductive,jin2013shared} applied structural regularizers across different feature views and tasks, and ensured the learned predictive models are similar for different tasks. 
However, these methods require multiple tasks directly sharing some label-dependent information with each other, which is only applicable to homogeneous tasks.
Focusing on heterogeneous tasks, many discriminative methods have been proposed, which map original heterogeneous features to a shared latent space through linear or nonlinear functions~\cite{zhang2011multi,jin2014multi,liu2018learning} or sparsity-driven feature selection~\cite{yang2009heterogeneous,jin2015heterogeneous}, and solve heterogeneous tasks jointly in the framework of discriminant analysis.
Generative models have achieve remarkable success in the past few years~\cite{wang2017topic,wang2018zero,wang2019improving}. However, to our knowledge, the generative solutions to heterogeneous multi-task learning have not been fully investigated.

\noindent\textbf{ICD code embedding and analysis of healthcare data}
Machine learning techniques have shown potential in many healthcare problems, {\em e.g.}, ICD code assignment~\cite{shi2017towards,baumel2017multi,mullenbach2018explainable,huang2018empirical}, admission prediction~\cite{ma2017dipole,liu2018early,xu2017patient}, mortality prediction~\cite{harutyunyan2017multitask,xu2018distilled}, procedure recommendation~\cite{mao2019medgcn}, medical topic modeling~\cite{choi2017gram,suo2018deep}, \emph{etc.}
Although these tasks have different objectives, they often share the same electronic health records data, {\em e.g.}, admission records. 
To learn multiple healthcare tasks jointly, various multi-task learning methods have been proposed~\cite{wang2014multi,alaa2017bayesian,suo2017multi,harutyunyan2017multitask,mao2019medgcn}.
Traditional multi-task learning methods imposed some structural regularizers on the features shared by different tasks~\cite{argyriou2007multi}. 
The work in~\cite{mao2019medgcn} applied GCNs~\cite{kipf2016semi} to extract features and jointly train models for medication recommendation and lab test imputation, which constitutes an attempt to apply GCNs to multi-task learning. 
However, introducing GCNs into the framework of generative heterogenous multi-task learning remains unexplored, that this paper seeks to address.

%------------------------------------------------------------------------
\section{Experiments}

% \begin{wraptable}{r}{0.4\textwidth}
\begin{table}[t]
	\centering
	\begin{threeparttable}[c]
	\begin{tabular}{
	    @{\hspace{4pt}}c@{\hspace{4pt}}
	    @{\hspace{4pt}}c@{\hspace{4pt}}
        @{\hspace{4pt}}c@{\hspace{4pt}}
        @{\hspace{4pt}}c@{\hspace{4pt}}
        @{\hspace{4pt}}c@{\hspace{4pt}}
        @{\hspace{4pt}}c@{\hspace{4pt}}
	}
	\hline\hline
	     &\textbf{Small} &\textbf{Median} & \textbf{Large} \\ \hline
		$|\mathcal{V}^d|$             & $247$     & $874$     & $2,765$ \\ 
		$|\mathcal{V}^p|$             & $75$      & $258$     & $819$ \\
		$|\mathcal{V}^a|$         & $28,315$  & $30,535$  & $31,213$  \\
	\hline\hline
	\end{tabular}
    \end{threeparttable}
    \caption{\small 
		Statistics of the MIMIC-III dataset.
	}\label{tab:dataset}
\end{table}
% \end{wraptable}

We test our method (GD-VAE) on the MIMIC-III dataset~\cite{johnson2016mimic}, which contains more than 58,000 hospital admissions with 14,567 disease ICD codes and 3,882 procedures ICD codes. 
For each admission, it consists of a set of disease and procedure ICD codes. 
Three subsets of the MIMIC-III dataset are considered, with summary statistics in Table~\ref{tab:dataset}. 
The subsets are generated by thresholding the frequency of ICD codes, {\em i.e.}, the ICD codes appearing at least 500/100/50 times and the corresponding non-empty admissions constitute the small/median/large subset.

To demonstrate the effectiveness of our method, we compare GD-VAE with state-of-the-art approaches on each of the healthcare tasks mentioned above.
Specifically, 
($i$) for topic modeling, we compare with LDA~\cite{blei2003latent}, AVITM~\cite{srivastava2017autoencoding} and BTM~\cite{yan2013biterm}. 
($ii$) For procedure recommendation, we compare with Bayesian Personalized Ranking (BPR)~\cite{rendle2009bpr},
Distilled Wasserstein Learning (DWL)~\cite{xu2018distilled}, and 
a VAE model designed for collaborative filtering (VAE-CF)~\cite{liang2018variational}.
We also compare with a baseline method based on Word2Vec~\cite{mikolov2013efficient}, which enumerates all possible disease-procedure pairs in each admission, and then recommends procedures according to the similarity between their embeddings and those of diseases.
($iii$) For admission-type prediction, we consider the following baselines: TF-IDF (combined with a linear classifier), Word2Vec (learning ICD code embeddings with Word2Vec~\cite{mikolov2013efficient}, and using the mean of the learned embeddings to predict the label), FastText~\cite{joulin2016bag}, SWEM~\cite{shen2018baseline} and LEAM~\cite{wang2018joint}. 
We use ``T'', ``R'' and ``P'' to denote topic modeling, procedure recommendation and admission-type prediction, respectively.
GD-VAE learns the three tasks jointly. 
To further verify the benefits of multi-task learning, we consider variations of our method that only learn one or two tasks, {\em e.g.}, GD-VAE (T) means only learning a topic model, and GD-VAE (TR) indicates the joint learning of topic modeling and procedure recommendation. 

\begin{table*}[t]
	\centering
	\caption{\small{Results on topic coherence for different models.} }\label{tab:TopicModel}
	\begin{threeparttable}
% 	\begin{small}
	\begin{tabular}{@{\hspace{1pt}}c@{\hspace{4pt}}
	c@{\hspace{8pt}}c@{\hspace{8pt}}c
	c@{\hspace{8pt}}c@{\hspace{8pt}}c
	c@{\hspace{8pt}}c@{\hspace{8pt}}c@{\hspace{1pt}}}
	\hline\hline
	    \multirow{2}{*}{Method} 
	    &\multicolumn{3}{c}{Small} 
	    &\multicolumn{3}{c}{Median} 
	    &\multicolumn{3}{c}{Large}\\
	   % \cline{2-10}
		&T=10 &T=30 &T=50 &T=10 &T=30 &T=50 &T=10 &T=30 &T=50 \\ 
		\hline
		           LDA~\cite{blei2003latent}       &0.110  &0.106  &0.098
	                         &0.123  &0.102  &0.107  
	                         &0.101  &0.106  &0.103 \\
	               AVITM~\cite{srivastava2017autoencoding}  &0.132  &0.125  & 0.121
	                         &0.135  & 0.110  &  0.107
	                         &0.123  & 0.116  & 0.108 \\ 
	               BTM~\cite{yan2013biterm}  & 0.117 & 0.109 & 0.105
	                         & 0.127 & 0.108  & 0.105 
	                         & 0.104 & 0.110  & 0.107 \\ \hline
		           GD-VAE (T)        & 0.142 & 0.141 & 0.135  
		                             & 0.140 & 0.137 & 0.132 
		                             & 0.128 & 0.129 & 0.123\\
		           GD-VAE (TP)       & 0.142 & 0.138 & 0.136 
		                             & 0.143 & 0.137 & 0.134 
		                             & 0.129 & 0.127 & 0.125 \\
		           GD-VAE (TR)       &0.147 & 0.147 & 0.144
		                             &0.146 & 0.141 & 0.137 
		                             &\textbf{0.136} & 0.133 & 0.127 \\
		           GD-VAE    & \textbf{0.151} & \textbf{0.149} & \textbf{0.145} 
		                     & \textbf{0.148} & \textbf{0.144} & \textbf{0.140} 
		                     & \textbf{0.136} & \textbf{0.137} & \textbf{0.131} \\
\hline\hline
	\end{tabular}
	\begin{tablenotes}
    \item \scriptsize{The standard deviation for GD-VAE and its variants is around 0.003.}
    \end{tablenotes}
    % \end{small}
    \end{threeparttable}
\end{table*}

\begin{table*}[ht]
	\centering
	\begin{threeparttable}
	\begin{small}
	\begin{tabular}{@{\hspace{1pt}}c@{\hspace{2pt}}@{\hspace{4pt}}c@{\hspace{4pt}}
	c@{\hspace{4pt}}c@{\hspace{4pt}}c
	c@{\hspace{4pt}}c@{\hspace{4pt}}c
	c@{\hspace{4pt}}c@{\hspace{4pt}}c
	c@{\hspace{4pt}}c@{\hspace{4pt}}c@{\hspace{2pt}}}
	\hline\hline
        \multirow{2}{*}{Dataset} 
	    &\multirow{2}{*}{Method} 
	    &\multicolumn{3}{c}{Top-1 (\%)} 
	    &\multicolumn{3}{c}{Top-3 (\%)} 
	    &\multicolumn{3}{c}{Top-5 (\%)} 
	    &\multicolumn{3}{c}{Top-10 (\%)}\\
	   % \cline{3-14}
		& &R &P &F1 &R &P &F1 &R &P &F1 &R &P &F1\\ 
		\hline
	               &Word2Vec~\cite{mikolov2013efficient} & 19.5 & 47.8 & 24.7
	                         & 35.4 & 34.9 & 30.8 
	                         & 47.1 & 29.6 & 32.0
	                         & 62.3 & 21.1 & 28.5\\
	               &DWL~\cite{xu2018distilled}      & 19.7 & 48.2 & 25.0
	                         & 35.9 & 35.2 & 31.3
	                         & 47.5 & 30.3 & 32.4
	                         & 63.0 & 20.9 & 28.7\\
		           &BPR~\cite{rendle2009bpr}      & 23.5 & 57.6 & 29.8
		                     & 44.8 & 43.5 & 38.7
		                     & 56.8 & 35.7 & 38.8 
		                     & 73.1 & 24.8 & 33.6\\
		  Small    & VAE-CF~\cite{liang2018variational}    & 24.0 & 57.8 & 30.7
		                     & 46.0 & 43.5 & 39.3 
		                     & 57.8 & 35.2 & 39.1 
		                     & 74.0 & 24.2 & 33.8 \\
		           &GD-VAE (R)    & 24.8 & 58.2  & 31.1
		                     & 46.5 & 43.4 & 39.5 
		                     & 58.1 & 35.3 & 39.2 
		                     & 74.5  & 24.4 & 34.0\\
		           &GD-VAE (RP)  & 25.0 & 58.3 & 31.3
		                     & 46.8& 43.5 & 39.5
		                     & 58.2& 35.4 & 39.2
		                     & 74.7& 24.5 & 34.1 \\
		           &GD-VAE (RT)   & 25.4 & 58.3 & 31.6
		                     & \textbf{47.0} & 43.6& 39.7
		                     & 58.5 & 35.9& 39.4
		                     & 75.2 & 24.8 & 34.3\\
		           &GD-VAE   & \textbf{25.6} & \textbf{58.6} & \textbf{31.8}  
		                     & \textbf{47.0} & \textbf{43.8} & \textbf{39.8} 
		                     & \textbf{58.7} & \textbf{36.2} & \textbf{39.6} 
		                     & \textbf{75.9} & \textbf{25.1} & \textbf{34.5} \\
		           \hline
		           &Word2Vec~\cite{mikolov2013efficient} & 7.8  & 27.6 & 10.9
		                     & 27.7 & 30.5 & 25.1
		                     & 38.3 & 26.9 & 27.7
		                     & 52.8 & 20.1 & 26.1\\
	               &DWL~\cite{xu2018distilled}      & 8.0  & 27.5 & 11.1
	                         & 27.9 & 30.8 & 25.2
	                         & 39.5 & 27.0 & 27.9
	                         & 53.9 & 20.9 & 27.4\\
		           &BPR~\cite{rendle2009bpr}      & 10.2 & 35.8 & 14.9
		                     & 38.6 & 40.2 & 34.3 
		                     & 49.3 & 33.3 & 34.9
		                     & 65.2 & 23.8 & 31.4\\
		  Median   & VAE-CF~\cite{liang2018variational}    & 21.2 & 52.9 & 26.2
		                     & 41.2 & 42.0 & 36.0 
		                     & 53.4 & 35.3 & 37.3 
		                     & 68.2 & 24.9 & 32.9 \\
		           &GD-VAE (R) & 22.0 & 55.1 & 27.9 
		                     & 42.3  & 41.2 & 37.2  
		                     & 54.0 & 35.7 & 37.8 
		                     & 69.3 & 25.2 & 33.1 \\
		           &GD-VAE (RP)   & 22.3 & 55.1  & 28.0 
		                     & 42.7 & 41.5 & 37.4 
		                     & 53.7 & 35.5 & 37.6
		                     & 69.6 & 25.1 & 33.4 \\
		           &GD-VAE (RT)      & 22.8 & 57.8 & 29.3 
		                     & 43.0 & 43.5 & 38.1  
		                     & 54.2 & 35.9 & 38.1
		                     & 70.1 & 25.2 & 33.6 \\
		           &GD-VAE     & \textbf{23.2} & \textbf{57.9} & \textbf{29.6} 
		                     & \textbf{43.2} & \textbf{43.9} & \textbf{38.2}
		                     & \textbf{54.6} & \textbf{36.0} & \textbf{38.4}
		                     & \textbf{70.4} & \textbf{25.3} & \textbf{33.7} \\
		           \hline
	               &Word2Vec~\cite{mikolov2013efficient} & 5.3  & 22.9 & 8.7 
	                         & 14.6 & 21.1 & 15.3 
	                         & 24.8 & 21.0 & 20.1 
	                         & 41.1 & 17.7 & 22.2\\
	               &DWL~\cite{xu2018distilled}      & 5.6  & 23.0 & 9.0 
	                         & 14.9 & 21.3 & 15.6
	                         & 24.8 & 21.4 & 20.5
	                         & 42.0 & 18.2 & 23.0\\
		           &BPR~\cite{rendle2009bpr}      & 7.3  & 26.7 & 10.2
		                     & 23.0 & 27.1 & 21.2 
		                     & 38.4 & 27.6 & 27.9
		                     & 56.6 & 21.7 & 28.0\\
		  Large    & VAE-CF~\cite{liang2018variational}    & 17.8 & 50.1 & 23.5
		                     & 35.2 & 37.9 & 33.4
		                     & 47.9 & 32.4 & 34.6
		                     & 63.0 & 21.7 &  30.2 \\
		          &GD-VAE (R) & 20.1 & 53.4 & 25.8
		                     & 37.2& 40.1 & 35.5 
		                     & 49.1& 32.5 & 35.2
		                     & 64.6 & 23.7 & 31.0\\
		           &GD-VAE (RP) & 20.4 & 53.3 & 26.1 
		                     & 37.9 & 39.7 & 35.9 
		                     & 49.9 & 32.7 & 35.5 
		                     & 65.1 & 24.0 & 31.2  \\
		           & GD-VAE (RT)  & 20.9 & 56.2 & 27.2  
		                     & \textbf{41.0} & 42.2 & 36.5
		                     & 50.9 & 35.1 & 36.6 
		                     & 66.0 & 24.7 & 32.5 \\
		           &GD-VAE   &\textbf{21.2} &\textbf{56.4} &\textbf{27.4} 
		                     &40.9 &\textbf{43.0} & \textbf{36.7} 
		                     &\textbf{51.4} &\textbf{35.2} &\textbf{36.8} 
		                     &\textbf{66.5} &\textbf{24.9} &\textbf{32.7} \\
	\hline\hline
	\end{tabular}
	\begin{tablenotes}
    \item \tiny{The standard deviation for GD-VAE and its variants is less than 0.2.}
    \end{tablenotes}
    \end{small}
    \end{threeparttable}
    \caption{\small{Comparison of various methods on procedure recommendation.}}\label{tab:recsys}
\end{table*}

\subsection{Configurations of Our Method}
We test various methods in 10 trials and record the mean value and standard deviation of the experimental results.
In each trial, we split the data into train, validation and test sets with a ratio of 0.6, 0.2 and 0.2, respectively. 
For the network architecture, we fix the embedding space to be $200$ for ICD codes and admissions, and a two-layer GCN~\cite{kipf2016semi} with residual connection is considered for the inference network. 
In terms of the dimension of latent variable, $\zv_T$ is identical to the number of topics for topic modeling and $200$ for the other two tasks, $\zv_R$ and $\zv_P$. 
In the aspect of the generative network, a linear layer is employed for both topic modeling and admission type prediction. 
For the procedure recommendation, a one-hidden layer MLP with $\tanh$ as the nonlinear activation function is used.
As for the hyper-parameters, we merge 10 randomly sampled admissions to generate a topic admission for our NBTM, such that $\yv_T$ is not too sparse, and $5,000$ samples are generated so as to train the model. Following~\cite{srivastava2017autoencoding}, the prior $\alphav$ is a vector with constant value 0.02.

\subsection{Topic modeling}\label{ssec:topic}
Topic coherence~\cite{mimno2011optimizing} is used to evaluate the performance of topic modeling methods. 
This metric is computed based on the normalized point-wise mutual information (NPMI), which has been proven to match well with human judgment~\cite{lau2014machine}.
Table~\ref{tab:TopicModel} compares different methods on the mean of NPMI over the top 5/10/15/20 topic words.
We find that LDA~\cite{blei2003latent} performs worse than neural topic models (including ours), which demonstrates the necessity of introducing powerful inference networks to infer the latent topics.
In terms of the GCN-based methods, GD-VAE and its variants capture global statistics between ICD codes and those between ICD codes and admissions, thus outperforming the three baselines by substantial margins. 

Compared with only performing topic modeling, {\em i.e.}, GD-VAE (T), considering more tasks brings improvements, and the proposed GD-VAE achieves the best performance. 
In terms of leveraging knowledge across tasks, we find that the improvements are largely contributed by procedure recommendation, and marginally from admission prediction. 
This is because procedure recommendation accounts for the concurrence between disease codes and procedure codes within an admission, while the topic model considers the concurrence between the codes from different admissions.
Both models capture the concurrence of ICD codes in different views, thus, naturally enhancing each other.

To further verify the quality of the learned topics, we visualize the top-5 ICD codes for some learned topics in the Supplementary Material. 
We find that the topic words are clinically-correlated.
For example, the ICD codes related to \emph{surgery} and those about \emph{urology} are concentrated in two topics, respectively. 
Additionally, each topic contains both disease codes and procedure codes, {\em e.g.,} ``d85306'' and ``p7817'' are \textit{orthopedic surgery} related disease and procedures, showing that disease and procedures can be closely correlated, which also implies the potential benefits brought to procedure recommendation.

\begin{table*}[t]
% \begin{wraptable}{r}{0.7\textwidth}
	\centering
	\caption{\small{Results on admission-type prediction. }}\label{tab:AdmTypePred}
    \resizebox{0.75\textwidth}{!}{%
	\begin{threeparttable}
	\begin{small}
	\begin{tabular}{@{\hspace{6pt}}c@{\hspace{4pt}}
	c@{\hspace{4pt}}c@{\hspace{4pt}}c
	c@{\hspace{4pt}}c@{\hspace{4pt}}c
	c@{\hspace{4pt}}c@{\hspace{4pt}}c@{\hspace{2pt}}}
	\hline\hline
	    Data 
	    &\multicolumn{3}{c}{Small} 
	    &\multicolumn{3}{c}{Median} 
	    &\multicolumn{3}{c}{Large}\\
	   % \cline{1-10}
		Method &P &R &F1 &P &R &F1 &P &R &F1 \\ 
		\hline
		           TF-IDF    &84.26  &87.19  &85.18 
	                         &86.12  &88.61  &87.22  
	                         &88.45  &89.10  &87.76 \\
	               Word2Vec~\cite{mikolov2013efficient}  & 85.08  & 87.89  &  86.23
	                         & 86.60 & 88.87  &  87.71
	                         & 87.11 & 89.16  &  88.12 \\
	               FastText~\cite{joulin2016bag}  &84.21  &87.15  &85.29 
	                         &86.66  &88.65  &87.39  
	                         &88.06  &89.23  &88.00 \\
		           SWEM~\cite{shen2018baseline}     &85.56  &88.10  &86.77
	                         &87.01  &89.28  &88.12  
	                         &87.55  &89.88  &88.67 \\
		           LEAM~\cite{wang2018joint}  & 85.34  & 88.03  & 86.55  
	                         & 87.03 & 89.29  &  88.14
	                         & 87.61 & 89.94  &  88.73\\ \hline
		           GD-VAE (P)        & 86.01 & 88.13  & 86.91
		                             &87.76 & 89.31 & 88.51
		                             &88.23 & 90.41 & 89.30 \\
		           GD-VAE (TP)       &86.18 & 88.52 & 87.22
		                             &87.82 & 89.21  & 88.52
		                             & 88.31& 90.56  & 89.41 \\
		           GD-VAE (RP)       & 86.87 & 89.38 &  87.93
		                             & 88.08&  89.57 &  88.82
		                             & 89.07&  90.98 & 90.00 \\
		           GD-VAE           & \textbf{87.00} & \textbf{89.60} & \textbf{88.01} 
		                            & \textbf{88.19} & \textbf{89.70} & \textbf{88.94}
		                            & \textbf{89.14} & \textbf{91.01} & \textbf{90.05} \\
\hline\hline
	\end{tabular}
    \begin{tablenotes}
    \item \tiny{The standard deviation for GD-VAE and its variants is around 0.05 on F1 score.}
    \end{tablenotes}
    \end{small}
    \end{threeparttable} }
\end{table*}
% \end{wraptable}

\begin{figure*}[t!]
	\subfigure[$t$-SNE visualization]{
		\includegraphics[width=0.235\textwidth, height=0.16\textwidth]{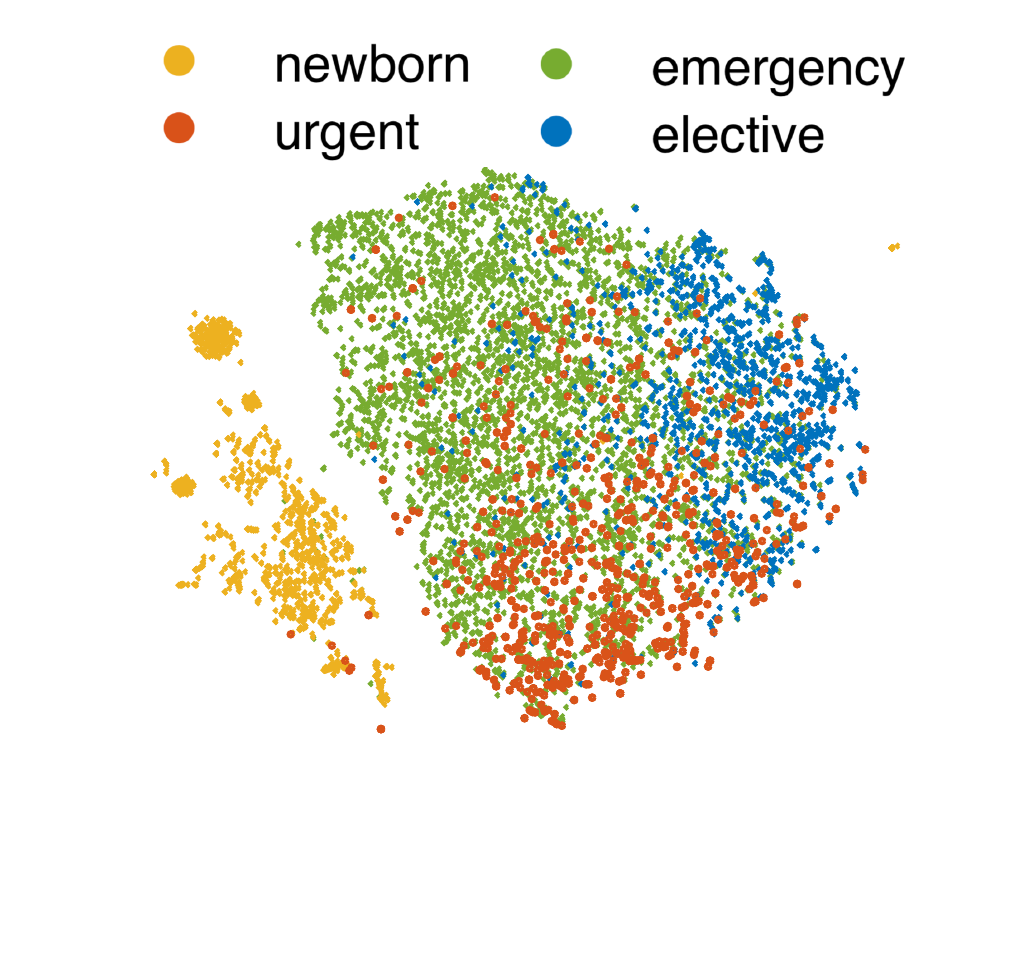}\label{fig:tSNE-ICD}
	}
	\subfigure[\scriptsize{Topic Modeling}]{
		\includegraphics[width=0.235\textwidth, height=0.16\textwidth]{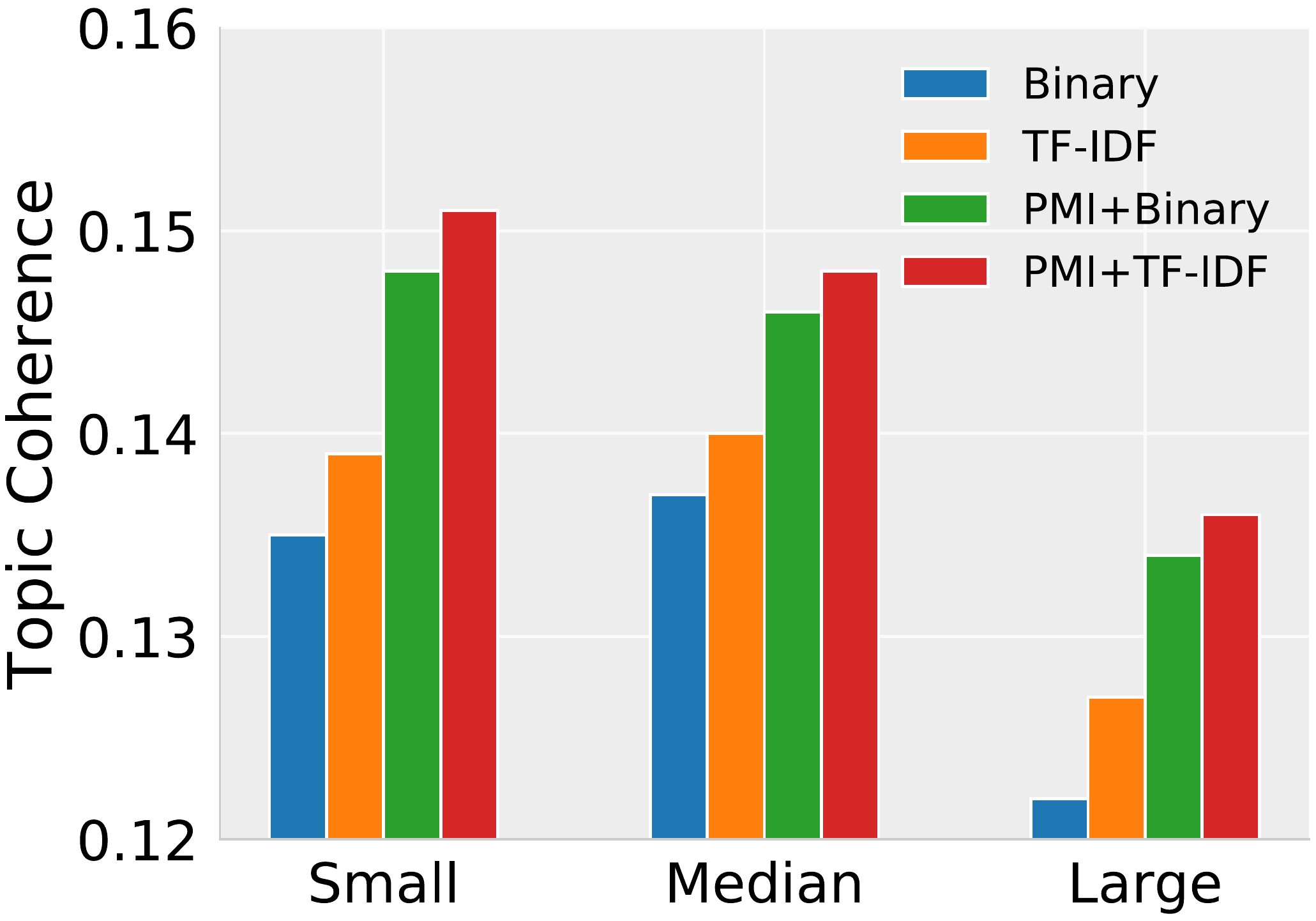}\label{fig:ablation_topic}
	}
	\subfigure[\scriptsize{Procedure Recommendation}]{
		\includegraphics[width=0.235\textwidth, height=0.16\textwidth]{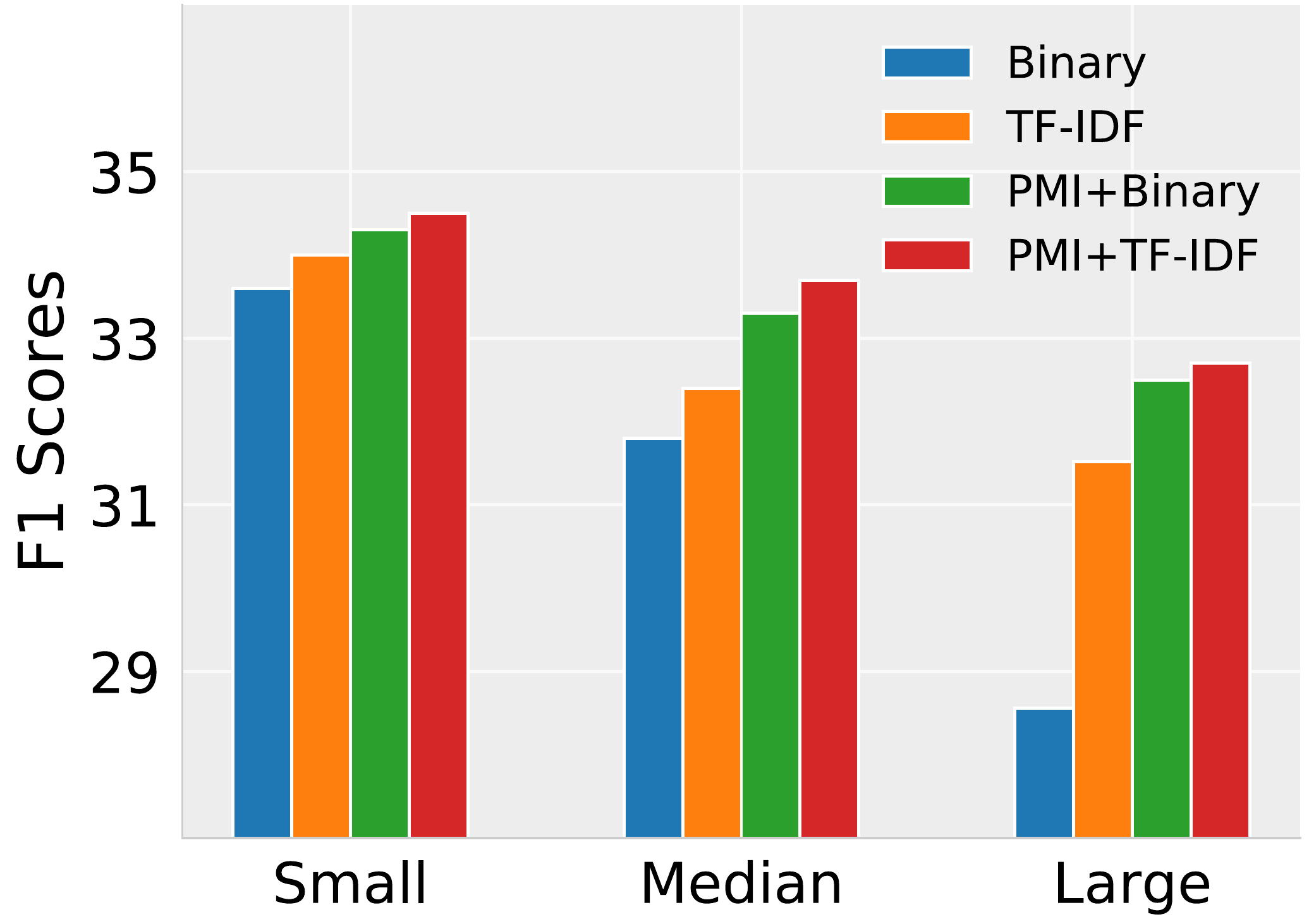}\label{fig:ablation_recsys}
	}
	\subfigure[\scriptsize{Admission-type Prediction}]{
		\includegraphics[width=0.235\textwidth, height=0.16\textwidth]{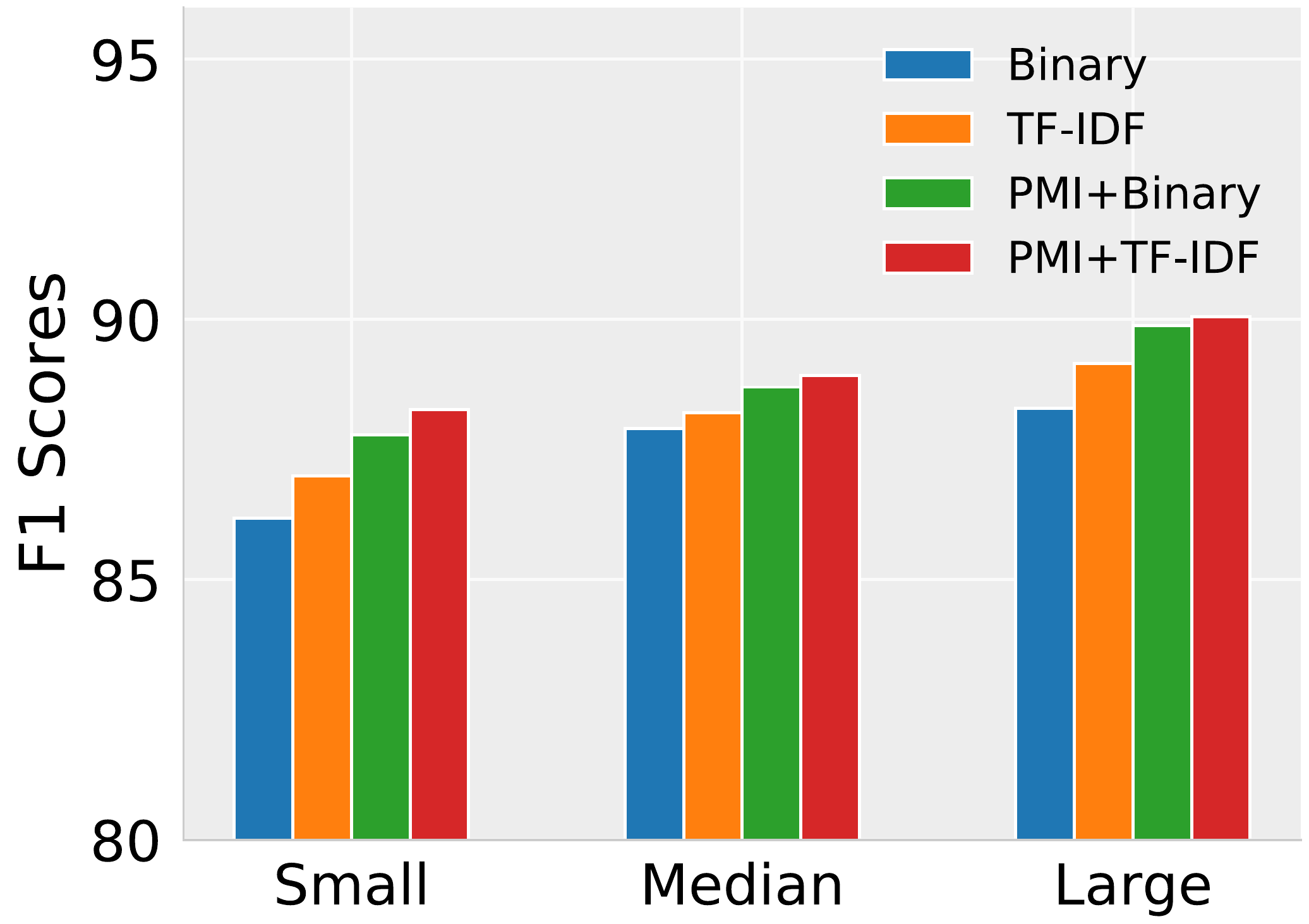}\label{fig:ablation_Prediction}
	}
	\caption{\small{(a) $t$-SNE visualization of $\zv_P$. (b, c, d) Rationality analysis of graph construction.}}
\end{figure*}

\subsection{Procedure recommendation}
Similar to~\cite{chen2018sequential,xu2018distilled}, we use top-$M$ precision, recall and F1-Score to evaluate the performance of procedure recommendation. 
Given the $n$-th admission, we denote $W_n$ and $Y_n$ as the top-$M$ list of recommended procedures and ground-truth procedures, respectively. 
The top-$M$ precision, recall and F1-score can be calculated as follows:
$P=\sum_{n=1}^N P_i=\sum_{n=1}^N\frac{|W_n\cap Y_n|}{|W_n|}$, $R=\sum_{n=1}^N R_i=\sum_{i=1}^N\frac{|W_n\cap Y_n|}{|Y_n|}$, $F1=\sum_{n=1}^N\frac{2P_n R_n}{P_n+R_n}$. 
Results are provided in Table~\ref{tab:recsys}.
GD-VAE (R) is comparable to previous state-of-the-art algorithms. 
With additional knowledge learned from topic modeling and admission-type prediction, the results can be further improved. 
Similar to the observation in the previous section, topic modeling contributes more to procedure recommendation than admission-type prediction, since both topic modeling and procedure recommendation explore the underlying relationship between diseases and procedures.

\subsection{Admission-type prediction}
Similar to procedure recommendation, we use precision, recall and F1-Score to evaluate the performance of admission-type prediction.
Results in Table~\ref{tab:AdmTypePred} show that GD-VAE outperforms its competitors. 
It is interesting to find that compared with topic modeling, procedure recommendation is more helpful to boost the results of admission-type prediction.
One possible explanation is that the admission type is more relevant to the set of procedures, hence the embedding jointly learned with procedure recommendation can better guide itself towards an accurate prediction, {\em e.g.,} it is likely to observe a {\em surgery} procedure in an {\em urgent} admission.
Additionally, to better understand the representation learned by GD-VAE, we visualize the inferred latent code $\zv_P$ with $t$-SNE~\cite{maaten2008visualizing}, as shown in Figure~\ref{fig:tSNE-ICD}. 
The $\zv_P$ is visually representative of different admission types being clustered into different groups.

\subsection{Rationality of graph construction}
To explore the rationality of our graph construction, we also compare the proposed admission graph with its variants. 
In particular, the proposed admission graph considers the PMI between ICD codes and the TF-IDF between ICD codes and admissions ({\em i.e.}, PMI + TF-IDF). 
Its variations include: ($i$) a simple graph with binary edges between admissions and ICD codes (Binary); 
($ii$) a graph only considering the TF-IDF between admissions and ICD codes (TF-IDF);
($iii$) a graph considering the PMI between ICD codes and the binary edges between admissions and ICD codes (PMI + Binary).
Results using different graphs are provided in Figures~\ref{fig:ablation_topic} through \ref{fig:ablation_Prediction}, which demonstrate that both PMI edges and TF-IDF edges make significant contributions to the performance of the proposed GD-VAE. 

%------------------------------------------------------------------------
\section{Conclusions}
We have proposed a graph-driven variational autoencoder (GD-VAE) to learn multiple heterogeneous tasks within a unified framework. 
This is achieved by formulating entities under different tasks as different types of nodes, and using a shared GCN-based inference network to leverage knowledge across all tasks.
Our model is general in that it can be easily extended to new tasks by specifying the corresponding generative processes. 
Comprehensive experiments on real-world healthcare datasets demonstrate that GD-VAE can better leverage information across tasks, and achieve state-of-the-art results on clinical topic modeling, procedure recommendation, and admission-type prediction.

%\noindent\textbf{Acknowledgement}: This research was supported in part by DARPA, DOE, NIH, ONR and NSF.
%------------------------------------------------------------------------
\clearpage
{
%\fontsize{7.25pt}{6.5}\selectfont
\small
\bibliographystyle{aaai} 
\bibliography{gvaehealth}
}

%------------------------------------------------------------------------
%\clearpage
%\vspace{10cm}
\begin{table*}[ht]
	\caption{Full description of topic words. }
	\centering
	\begin{adjustbox}{max width=\textwidth}
		\setlength{\tabcolsep}{5pt}
		\begin{tabular}{@{\hspace{1pt}}c@{\hspace{6pt}}
				@{\hspace{6pt}}l@{\hspace{1pt}}
				@{\hspace{6pt}}l@{\hspace{1pt}}} 
			\hline\hline
		     & ICD codes & Description \\ \hline 
		     \multirow{5}{*}{Topic 1} & d8708 & Other Specified open wounds of ocular adnexa \\ 
		     &d85306 & Other and unspecified intractranial hemorrhage following injury \\ 
		     &dE8192 & Closed reduction of mandibular fracture \\
		     &p7817 & Application of extrenal fixator device, tibia and fibula \\
		     &p2751 & Suture of Iaceration of lip \\ \hline 
		     \multirow{5}{*}{Topic 2} &p3783 & Initial insertion of dual-chamber device \\
		     &p3764 & Removal of external heart assist system or device \\
		     &d7660 & Exceptionally large baby\\
		     &93514 & Open heart valvuloplasty of tricuspid value without replacement\\
		     &d41021 & Acute myocardial infarction of inferolateral wall, initial episode of care\\ \hline 
		     \multirow{5}{*}{Topic 3} &d8774 & Retrograde pyelogram \\
		     &d5503 & Percutaneous nephrostomy without gragmentation \\
		     &d6084 & Other inflammatory disorders of male genital organs \\
		     &p560 &  Transurethral removal of obstruction from ureter and renal pelvis\\
		     &d1981 & Secondary malignant neoplasm of other urinary organs \\ \hline 
		     \multirow{5}{*}{Topic 4} &p3951 & Clipping of aneurysm \\
		     &d2242 & Frontal sunusectomy \\
		     &p109 & Other cranial puncture \\
		     &d78552 & Other craniotomy \\
		     &d51883 & Other specified acquired deformity of head \\ \hline 
		     \multirow{5}{*}{Topic 5} &d33520 & Amyotrophic lateral sclerosis \\
		     &d51902 & Mechanical complication of tracheostomy \\
		     &p3199 & Other operations on trachea \\
		     &d7708 & Other tracheostomy complications \\
		     &d8718 & Chronic respiratory failure \\ \hline 
		     \multirow{5}{*}{Topic 6} &d7783 & Other hypothermia of newborn \\
		     &p640 & Circumcision \\
		     &d76406 & ``light-for-dates'' without mention of fetal malnutrition\\
		     &d7731 & Hemolytic disease of fetus or newborn due to ABO isoimmunization\\
		     &p9983 &  Other phototherapy\\ \hline 
		     \multirow{5}{*}{Topic 7} &d45620 & Esophageal varices in diseases classified elsewhere, with bleeding \\
		     &p9635 & Gastric Gavage \\
		     &d4560 & Esophageal variaces with bleeding\\
		     &d4233 &  Endoscopic excision or destruction of lession or tissue of esophagus \\
		     &d53240 & Chronic or unspecified duodenal ulcer with hemorrhage, without mention of obstruction \\ 
			\hline\hline
		\end{tabular}\label{tab_topic_description}
		\end{adjustbox}
\end{table*}

\section{GCN-based Inference Network}
Graph convolutional network (GCN)~\cite{kipf2016semi} has attracted much attention for leveraging information representations for nodes and edges, and is promising to tasks with complex relational information~\cite{kipf2016semi,hamilton2017inductive,yao2018graph}.

Given a graph $G(\mathcal{V}, \mathcal{X}, \Av)$, 
a graph convolution layer derives the node embeddings $\Hv\in\mathbb{R}^{N\times \Tilde{d}}$ via 
\begin{align}
    \Hv = \text{ReLU} (\Tilde{\Av}\mathcal{X}\Wv)\,, \nonumber
\end{align}
where $\Tilde{d}$ is the dimension of feature space after GCN, $\Tilde{\Av} = \Dv^{-\frac{1}{2}}\Av\Dv^{-\frac{1}{2}}$ is the normalized version of adjacency matrix, $\Dv$ is a diagonal degree matrix with $\Dv_{ii} = \sum_j \Tilde{\Av}_{ij}$, and $\Wv$ is a matrix of trainable graph convolution parameters. 
The GCN aggregates node information in local neighborhoods to extract local substructure information. 
In order to incorporate high-order neighborhoods, we can stack multiple graph convolution layers as 
\begin{align}
    \Hv^{l+1} = \text{ReLU}(\Tilde{\Av}\Hv^l \Wv^l)\,, \nonumber 
\end{align} 
where $\Hv^0 = \Xv$, $\Hv^l$ is the output of the $l$-th graph convolutional layer. However, GCN can be interpreted as Laplacian smoothing. 
Repeatedly applying Laplacian smoothing may fuse the features over vertices and make them indistinguishable~\cite{li2018deeper}. 
Inspired by~\cite{he2016deep}, we alleviate this problem in our inference network by adding shortcut connections between different layers. 

\section{Description of topic ICD codes}
A full description of the top-5 topic ICD codes is shown in Table~\ref{tab_topic_description}.

\end{document}